\documentclass[conference]{IEEEtran}

\IEEEoverridecommandlockouts \IEEEpubid{\makebox[\columnwidth]{ 979-8-3503-0737-5/23/\$31.00~\copyright2023 IEEE \hfill} \hspace{\columnsep}\makebox[\columnwidth]{ }}

\IEEEspecialpapernotice{Wild-and-Crazy-Idea Paper}

\usepackage{cite}
\usepackage{amsmath,amssymb,amsfonts}
\usepackage{algorithmic}
\usepackage{graphicx}
\usepackage{textcomp}
\usepackage[dvipsnames]{xcolor}

\usepackage{soul}
\usepackage[normalem]{ulem}
\usepackage{etoolbox}
\soulregister\cite7
\soulregister\ref7
\soulregister\pageref7
\soulregister\gls7
\soulregister\glspl7
\let\origcite\cite
\renewcommand\cite[1]{\mbox{\origcite{#1}}}
\newtoggle{ShowChanges}
\togglefalse{ShowChanges}
\newcommand{\add}[1]{\iftoggle{ShowChanges}{{\protect\textcolor{blue}{\uwave{#1}}}}{#1}}
\newcommand{\remove}[1]{\iftoggle{ShowChanges}{{\protect\textcolor{red}{\st{#1}}}}{}}
\newcommand{\replace}[2]{\remove{#1}\add{#2}}

\def\BibTeX{{\rm B\kern-.05em{\sc i\kern-.025em b}\kern-.08em
    T\kern-.1667em\lower.7ex\hbox{E}\kern-.125emX}}

\usepackage{xspace}
\makeatletter
\DeclareRobustCommand\onedot{\futurelet\@let@token\@onedot}
\def\@onedot{\ifx\@let@token.\else.\null\fi\xspace}

\def\eg{\emph{e.g}\onedot} 
\def\ie{\emph{i.e}\onedot}

\def\etal{\emph{et al}\onedot}
\makeatother

\makeatletter
\def\ps@IEEEtitlepagestyle{%
  \def\@oddfoot{\mycopyrightnotice}%
}
\def\mycopyrightnotice{%
\begin{minipage}{\textwidth}
\centering \footnotesize
Copyright~\copyright~2023 IEEE. Personal use of this material is permitted.  Permission from IEEE must be obtained for all other uses, in any current or future media, including reprinting/republishing this material for advertising or promotional purposes, creating new collective works, for resale or redistribution to servers or lists, or reuse of any copyrighted component of this work in other works.
\end{minipage}
}
\makeatother

\usepackage{hyperref}

\begin{document}

\iftoggle{ShowChanges}{\input{srcs/rebuttal_summary}}{}


\bstctlcite{IEEEexample:BSTcontrol}

\title{Neuro-symbolic Empowered Denoising Diffusion Probabilistic Models for Real-time Anomaly Detection in Industry 4.0%
\thanks{This study was carried out within the PNRR research activities of the consortium iNEST (Interconnected North-Est Innovation Ecosystem) funded by the European Union Next-GenerationEU (Piano Nazionale di Ripresa e Resilienza (PNRR) – Missione 4 Componente 2, Investimento 1.5 – D.D. 1058  23/06/2022, ECS\_00000043). This manuscript reflects only the Authors’ views and opinions, neither the European Union nor the European Commission can be considered responsible for them.}
}

\author{%
\IEEEauthorblockN{%
Luigi Capogrosso\IEEEauthorrefmark{1}, %
Alessio Mascolini\IEEEauthorrefmark{2}, %
Federico Girella\IEEEauthorrefmark{1}, %
Geri Skenderi\IEEEauthorrefmark{1}, %
Sebastiano Gaiardelli\IEEEauthorrefmark{1},\\%
Nicola Dall'Ora\IEEEauthorrefmark{1}, %
Francesco Ponzio\IEEEauthorrefmark{3}, %
Enrico Fraccaroli\IEEEauthorrefmark{4}, %
Santa Di Cataldo\IEEEauthorrefmark{2}, %
Sara Vinco\IEEEauthorrefmark{2},\\%
Enrico Macii\IEEEauthorrefmark{3}, %
Franco Fummi\IEEEauthorrefmark{1}, %
Marco Cristani\IEEEauthorrefmark{1}%
}%
\IEEEauthorblockA{\IEEEauthorrefmark{1}{\emph{Dept. of Engineering for Innovation Medicine}, University of Verona, Italy, \tt name.surname@univr.it}}%
\IEEEauthorblockA{\IEEEauthorrefmark{2}{\emph{Dept. of Control and Computer Engineering}, Polytechnic of Turin, Italy, \tt name.surname@polito.it}}%
\IEEEauthorblockA{\IEEEauthorrefmark{3}{\emph{Dept. of Regional and Urban Studies and Planning}, Polytechnic of Turin, Italy, \tt name.surname@polito.it}}%
\IEEEauthorblockA{\IEEEauthorrefmark{4}{\emph{Dept. of Computer Science}, University of North Carolina at Chapel Hill, USA, \tt enrifrac@cs.unc.edu}}%
}

\maketitle
\IEEEpubidadjcol

\begin{abstract}
Industry 4.0 involves the integration of digital technologies, such as IoT, Big Data, and AI, into manufacturing and industrial processes to increase efficiency and productivity. 
As these technologies become more interconnected and interdependent, Industry 4.0 systems become more complex, which brings the difficulty of identifying and stopping anomalies that may cause disturbances in the manufacturing \replace{procedure}{process}. 
This paper aims to propose a diffusion-based model for real-time anomaly prediction in Industry 4.0 processes. 
Using a neuro-symbolic approach, we integrate industrial ontologies in the model, thereby adding formal knowledge on smart manufacturing. 
Finally, we propose a \replace{new}{simple yet effective} way of distilling diffusion models through Random Fourier Features for deployment on an embedded system for direct integration into the manufacturing process. 
To the best of our knowledge, this approach has never been explored before.
\end{abstract}


\section{Context and motivation}
\label{sec:introduction}

The dawn of Industry 4.0 has \replace{brought about}{has ushered} a digital revolution in industrial processes, significantly increasing productivity, efficiency, and quality. 
This transformation is facilitated by the Industrial Internet of Things (IIoT)~\cite{sisinni2018industrial}, which involves the interconnection of industrial devices, equipment, and systems through the Internet. 
The generation and collection of massive amounts of diverse data enabled by IIoT provide insights into various aspects of production, such as process optimization, quality control, and resource allocation. But, on the other hand, bring new challenges in data analysis and management. 

One of the most critical tasks in this context is \textit{anomaly detection}, which refers to the identification of irregular events or behaviors from the data. 
Malfunctioning equipment, human errors, external disturbances, or unforeseen circumstances can disrupt production, pose safety hazards, or lower product quality~\cite{chandola2009anomaly}. 
As such, the development of techniques able to accurately and timely detect anomalies is crucial for maintaining smooth and efficient operations, as well as ensuring the safety and well-being of workers.

Although remarkable improvements have been made in recent literature, especially in the field of deep learning \remove{methods}, existing anomaly detection approaches \replace{fall short of}{still struggle to} effectively \replace{modeling}{model} the behavior of complex machinery while maintaining their computational effort within feasible limits for real-time execution at the edge~\cite{pang2021deep}. 
This motivation is the core of our research proposal.

\section{Our Proposal}
\begin{figure}[t]
\centering
\includegraphics[width=\columnwidth]{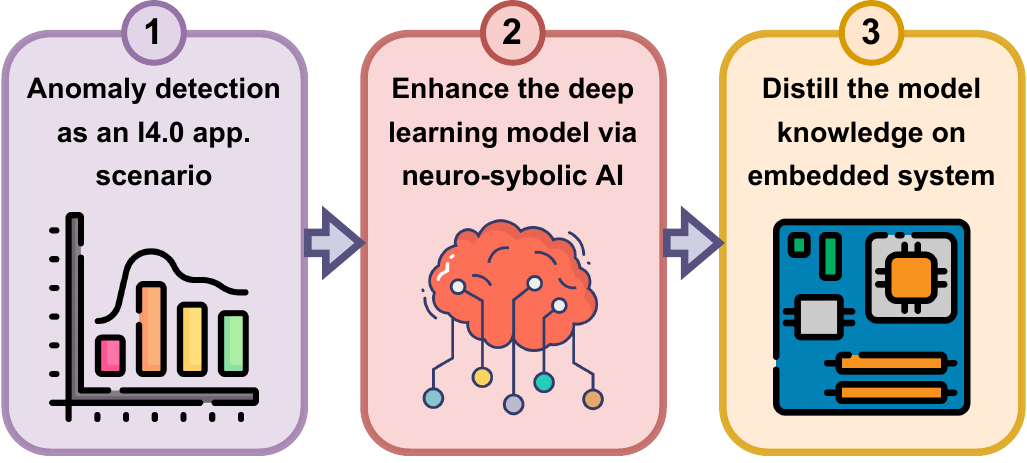}
\caption{The key topics of our proposal. \textit{1)} Starting from the open problem of anomaly detection in an Industry 4.0 scenario, \textit{2)} we propose to use a recent category of deep learning models (\ie{}, diffusion models) to address this problem, by enhancing them with neuro-symbolic learning, which is one of the trending topics of the moment in the field of AI. \textit{3)} Finally, we propose a distillation strategy to transfer the knowledge into an embedded system for real-time usage on the product line.}
\label{fig:teaser}
\end{figure}
\remove{Imagine a manufacturing plant that produces digital circuits disseminated with several heterogeneous sensors. Now, let us focus on a specific part of the product line, \ie{}, the one in which a robotic arm moves the digital board from the assembly to the testing line.} 
We want to enhance the productivity and safety of \replace{this}{the} production line by implementing a system that can effectively detect irregular behaviors of the \replace{robotic arm}{system}, even when human attention is lacking or unavailable. 
Traditional methods for achieving this goal rely on purely symbolic approaches, as companies need assurance that the system will provide explainable predictions and will not demand extensive computing resources. 
However, due to the simplicity of these methods, they are often limited in their ability to recognize anomalies that rely on context or relationships between sensors. 
Another option is to rely on deep learning models, which can learn complex relationships in the sensed data. 
Unfortunately, they are often avoided due to their ``black-box'' nature and the amount of effort required to run even the smallest models in real-time. 
Furthermore, they might require more expensive hardware and higher power usage, without the guarantee to match or exceed the performance of symbolic approaches.

To address these concerns, \replace{we propose a new anomaly detection approach that combines \textit{diffusion models}, \textit{neuro-symbolic AI}, and \textit{knowledge distillation} via teacher-student modeling.}{this paper presents an innovative approach to the integration of a formally constrained diffusion model for anomaly detection with embedded systems for industrial applications, merging neuro-symbolic methods, formal constraints to enhance reliability and safety in industrial systems, and knowledge distillation.}
This innovative solution strives to improve monitoring capabilities without sacrificing the reliability and efficiency required in an Industry 4.0 production line setting.
Figure~\ref{fig:teaser} shows our proposed anomaly detection flow.
What follows is a detailed explanation of the flow which, we hope, might pose the basis for further investigation and research in this area.

We approach anomaly detection as a task of Out-Of-Distribution (OOD) classification. 
Anomaly detection is the task of identifying examples in a dataset that are different or unusual compared to most other examples. 
Similarly, OOD classification is the task of identifying examples that do not belong to a known data distribution. 
Both tasks involve identifying examples dissimilar to the majority of the other examples, and both require the ability to identify patterns in the data that are significantly different from the norm. 
Inspired by the work of~\cite{graham2022denoising}, we propose a Denoising Diffusion Probabilistic Model (DDPM)~\cite{ho2020denoising} as the backbone for our anomaly detection process. 
We take advantage of DDPMs' ability to understand a given distribution latent structure to detect OOD samples in our data via differences in signal reconstruction. \replace{By doing so, we can assign a label (in or out of distribution) to each of our training samples in a completely unsupervised way.}{The final goal of the DDPM will be to label the training data in a fully unsupervised way.}

Our second contribution is exploiting ontologies to add formal and additional knowledge as properties that can be integrated within the deep learning model\add{, to constrain the diffusion model to learn a data distribution that always respects the given logical axioms}. We plan to do so by extending the methodology presented in~\cite{kocher2020formal}, where authors present a formal model able to represent abstract capabilities, structure, and executable skills of a manufacturing system through so-called ontology design patterns based on industrial standards, \eg{}, DIN 8580 \add{\cite{DIN8580}}, VDI 3682\add{\cite{VDI3682}}, and VDI 2860\add{\cite{VDI2860}}. 
By enriching our network with domain-specific knowledge, we ensure the model starts from a formal understanding of the data and can detect the anomalies that a purely symbolic system should be able to detect. 
Combined with the Deep Generative Model (DGM), this allows the network to learn more complex rules and relationships, and thus more sophisticated anomalies.

Specifically, we plan to use a neural-symbolic AI~\cite{sarker2021neuro} approach to support the learning phase of neural networks using the satisfaction of a first-order logic knowledge base as an objective\add{, similarly to~\cite{roychowdhury2021regularizing}}. 
In our case, this is provided by the ontologies that formalize the smart manufacturing knowledge in an interoperable way. 
From the formal conceptualization given by the ontologies, we obtain the first-order logic knowledge base containing a set of axioms. 
At this point, we have some predicates, or functions appearing in these axioms that we want to learn, and some data available that we can use to learn the parameters of those symbols. 
The idea is to embed the logical axioms into the loss function of our diffusion model. 
The goal of our model then becomes finding solutions in the hypothesis space that maximally satisfy all the axioms contained in our knowledge base.

This hybrid method, which combines both data-driven and knowledge-based techniques, allows the neural network to discern the remaining nuances and intricacies of the system, which should ultimately enhance its overall performance and reliability. 
The described system is explainable by design, as it allows the user to know which sensors are reporting anomalous data, and which values the system would have considered acceptable.

Finally, we \replace{tackle}{address} the \replace{problem}{challenge} of running the algorithm in real-time on the product line, which is in direct contrast with the computationally intensive nature of deep learning models\replace{and, in particular,}{, particularly} diffusion models. 
In order to solve this, our third proposal consists of the use of Random Fourier Features (RFF)~\cite{rahimi2007random} to train a classifier on the \add{binary }labels obtained by our DDPM, effectively distilling its knowledge\add{, without} using a teacher-student learning paradigm~\cite{gou2021knowledge}, into a more lightweight and practically useful detector. 
\remove{Specifically,}RFF-based classifiers represent an extension of linear classifiers that project the data into a higher dimensional space, where the classes are more easily separable, in order to model arbitrary non-linear functions. \add{By training a RFF model on the pseudo-labels created by the NeSy-DDPM, it is plausible to think that RFFs can be used to provide an optimal decision boundary in kernel space~\cite{nguyen2019scalable} for our proposed anomaly detection classifier. To sum up, the RFF must not capture the posterior inferred by the diffusion model per se, but they must be able to define features that support the (binary) label distribution generated by the diffusion-based-OOD~\cite{graham2022denoising}.}
Given a kernel function $K(x, y)$, we can represent it as an inner product in a higher-dimensional space via the feature map:
\begin{equation}\label{eq:rff}
\small
\phi(x) = \begin{bmatrix}
\cos(w_1^Tx + b_1) \\
\sin(w_1^Tx + b_1) \\
\vdots \\
\cos(w_D^Tx + b_D) \\
\sin(w_D^Tx + b_D) \\
\end{bmatrix}\;,
\end{equation}
where $D$ is the desired dimensionality of the feature space, and $w_i$ and $b_i$ are randomly generated parameters.

To train a linear classifier using RFF, we first compute the feature representation for each input data point using the above formula and then use it to predict the label previously assigned to that sample by our DDPM approach.
Once the linear classifier is trained on these features, we can make predictions on new data points by computing their feature representation and then multiplying them by the learned weight vector $w$ and adding a bias term $b$:
\begin{equation}\label{eq:inference}
p(y_{pred}) = \frac{1} {1 + e^{-(w^T \phi(x_{new}) + b)}}\;.
\end{equation}
\remove{As one can easily see} \replace{from}{From}~\ref{eq:inference}, the entire inference process requires only two matrix multiplications, making them extremely lightweight and quickly executable on low-end hardware.

\begin{figure*}[t]
\centering
\includegraphics[width=\linewidth]{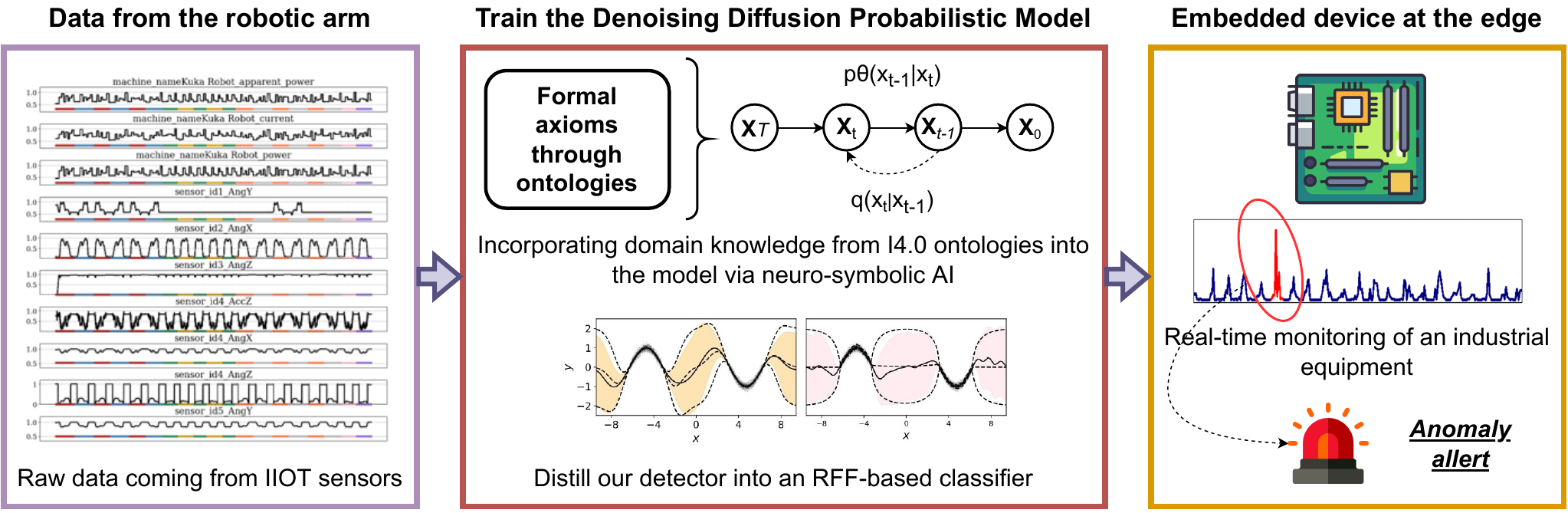}
\caption{Detailed overview of our proposal. The time series on which we want to train the diffusion model are acquired by the sensors of the robotic arm. \replace{The diffusion model represents the latest advancement in deep learning, and thus, potentially better performance compared with other previous models}{Diffusion models represent the latest advancement in deep learning, and thus, are potentially capable of better performance compared to previous models}. This model is enhanced through a neural-symbolic approach using the satisfaction of a first-order logic knowledge base as an objective, which, in our case, is provided by the ontologies that formalize the smart manufacturing knowledge in an interoperable way. Finally, we face the problem of running the algorithm in real-time on the product line effectively distilling its knowledge into an RFF-based classifier. As a result, the entire inference process requires only two matrix multiplications, which can be executed quickly on resource-constrained devices.}
\vspace{-1.0em}
\label{fig:proposal}
\end{figure*}
In conclusion, our idea, shown in Figure~\ref{fig:proposal}, consists of a series of proposals in order to leverage the power of diffusion models for anomaly detection from time series data. 
The framework can capture the complex spatiotemporal dependencies between the sensors, handle large amounts of data in a rigorous way, and enable proactive anomaly detection and mitigation, thereby enhancing the reliability, safety, and efficiency of industrial processes. 
For all of these reasons, we think that our proposal can be a valuable tool for Industry 4.0.
\section{Related Work}
\label{sec:related}
\replace{This section provides details on the techniques our proposal relies upon.}{This section introduces the techniques we rely upon.}

\subsection{Diffusion models}
Given observed samples $x$ from a distribution of interest, the goal of a generative model is to learn to estimate its true data distribution $p(x)$.
Once learned, we are able to use the learned model to evaluate the likelihood of newly observed data.
DDPMs are parameterized Markov chains models, able to learn the latent structure of data by modeling the way data points diffuse through the latent space.
They emerged in 2015~\cite{sohl2015deep} as a powerful new family of deep generative models with a record-breaking performance in a broad range of applications, spanning from image synthesis and multi-modal modeling to temporal analysis and natural language problem~\cite{yang2022diffusion}.
Referring to our problem, there are three principal works related to OOD with diffusion models.

The authors of~\cite{graham2022denoising} utilize DDPMs as denoising autoencoders, where the amount of noise applied externally controls the \replace{bottleneck}{strength of the conditioning}.
They suggest employing DDPMs to reconstruct a noised input across various noise levels and leverage the resulting multidimensional reconstruction error to classify OOD inputs.
The experiments demonstrate that the proposed DDPM-based technique surpasses both reconstruction-based methods and state-of-the-art generative approaches.

\remove{In contrast, in~\cite{fang2022out}, the authors explore the generalization of OOD detection by delving into the probably approximately correct (PAC) learning theory of OOD detection, which was proposed as an open problem by researchers.
Initially, the authors establish a necessary condition for OOD detection learnability.
They then employ this condition to demonstrate several impossibility theorems for OOD detection learnability in specific scenarios.
Although these theorems are discouraging, the authors discover that some of the conditions of the theorems may not apply in practical situations. 
Consequently, they provide several necessary and sufficient conditions to describe the learnability of OOD detection in practical scenarios.} 

As previously stated, DGMs appear to be a natural choice for detecting OOD inputs.
However, these models have been observed to assign higher probabilities or densities to OOD images than images from the training distribution.
The work in~\cite{zhang2021understanding} addresses this behavior and attributes it to model misestimation, suggesting it to be a more probable cause, rather than the misalignment between likelihood-based OOD detection and our distributions of interest.
They also demonstrate how even slight estimation errors can result in OOD detection failures, which carries implications for future research in deep generative modeling and OOD detection.
This work, although primarily focused on image inputs, has shown that DDPMs can be utilized to address this problem.
This provides theoretical foundations for our idea, even though it may seem unconventional.

\subsection{Neuro-symbolic AI}
Neuro-symbolic AI refers to the combination of artificial neural networks with symbolic knowledge representation and reasoning techniques used in symbolic AI.
This approach aims to overcome the limitations of traditional rule-based symbolic AI by incorporating both logical reasoning and statistical inference into the same model~\cite{sarker2021neuro}.
\remove{Hence, it is seen as a promising approach to solving complex problems that require both explicit rule-based reasoning and learning from data, such as natural language understanding, robotics, and scientific discovery~\cite{hitzler2022neuro}.}
\remove{Future research in this area is expected to lead to significant advancements in intelligent systems, autonomous robotics, and personalized medicine.}

One of these studies is that of Zheng \etal{}~\cite{zheng2022jarvis}, tackling the challenge of developing a conversational embodied agent capable of executing real-life tasks.
\remove{Such an agent requires effective human-agent communication, multi-modal understanding, and long-range sequential decision-making, among other capabilities.}
Traditional symbolic methods suffer from scaling and generalization issues, while end-to-end deep learning models face data scarcity and high task complexity, and are often difficult to interpret.
To take advantage of both approaches, the authors propose a neuro-symbolic commonsense reasoning framework for generalizable and interpretable conversational embodied agents.

Another recent work is the one by Huang \etal{}~\cite{huang2023laser}, in which they propose a neuro-symbolic approach that learns semantic representations by leveraging logic specifications that can capture rich spatial and temporal properties in video data.
\remove{In particular, they formulate the problem in terms of alignment between raw videos and specifications.}

Furthermore, Siyaev \etal{}~\cite{siyaev2023interaction} proposed a neuro-symbolic reasoning approach for interacting with 3D digital twins through natural language.
This method can comprehend user requests and contexts to manipulate 3D components of digital twins and execute installations and removal procedures independently by reading maintenance manuals.

Overall, all these works demonstrate the potential of neuro-symbolic AI in various domains\remove{, highlighting its ability to combine the strengths of symbolic reasoning and deep learning to create more robust and interpretable models}.
The proposed approaches not only improve the accuracy of predictions but also provide valuable insights into the decision-making process, making them more transparent and trustworthy.
\section{Discussion}
\label{sec:discussion}

This paper outlines a novel idea that combines neuro-symbolic diffusion models and RFF for real-time anomaly detection. 
This section discusses the reasons why this idea is considered wild and crazy and the potential implications of successfully implementing such a model.

First and foremost, while the idea of integrating symbolic and neural systems has been explored in other contexts, such as in hybrid neuro-symbolic models, the concept of neuro-symbolic diffusion models is entirely new and has never been attempted before.
This approach proposes to combine the strengths of symbolic and neural systems in a single model, potentially enabling a more robust representation of complex information.
\replace{This represents a significant leap in the field and could lead to unforeseen challenges and complications.
However, if successful, this novel approach could also open up new possibilities for more efficient and effective learning models.}{At the same time, it relies on domain-specific logical constraints to guide the DDPM during training, meaning that careful considerations on the formalisms used are needed by domain experts.}

\replace{Secondly, using RRF for distilling neural networks is also an entirely new concept. RFF has been used in other contexts to approximate kernel functions and improve the efficiency of specific learning algorithms.
Still, their application to neural network distillation is unexplored and presents numerous potential challenges.
Despite these challenges, integrating RFF in neural network distillation offers a promising avenue for improving the efficiency and scalability of deep learning models.}
{Secondly, while RFF classifiers have been used in other contexts, utilizing them as a means to distill DDPMs is unexplored and presents numerous challenges. That being said, their low computation complexity offers a promising avenue for deployment on embedded systems, allowing us to utilize the power of deep learning models even in resource-constrained environments.}
\remove{If successful, this approach could lead to significant advancements in deploying neural networks in resource-constrained environments and enable more widespread adoption of deep learning techniques in industrial domains.}

The complexity of Industry 4.0 systems and the datasets they produce can present significant challenges for any anomaly detection approach.
Anomaly detection is a critical task that demands high accuracy and reliability; as such, any approach utilized for this purpose must undergo thorough testing and validation to ensure its effectiveness.
It is vital to consider whether the proposed approach can handle the complexity of these systems and datasets and whether it can scale to meet the demands of real-world applications. 

Although it may seem unconventional at first glance, our approach is grounded in sound scientific principles and has the potential to significantly contribute to the field of smart manufacturing, offering a promising solution to the challenges posed by real-time anomaly detection in Industry 4.0.

\bibliographystyle{IEEEtran}
\bibliography{bibi}

\end{document}